\pdfoutput=1

\documentclass[11pt]{article}

\usepackage[final]{acl}

\usepackage{times}
\usepackage{latexsym}

\usepackage[T1]{fontenc}

\usepackage[utf8]{inputenc}

\usepackage{microtype}

\usepackage{hyperref}       
\usepackage{url}            
\usepackage{inconsolata}
\usepackage{graphicx}
\usepackage{amsfonts}
\usepackage{amsmath}
\usepackage{amssymb}
\usepackage{mathtools}
\usepackage{bm}
\usepackage{bbm}
\usepackage{booktabs}
\usepackage{multirow}
\usepackage{multicol}
\usepackage{nicefrac}       
\usepackage{microtype}      
\usepackage{xcolor}         
\usepackage{adjustbox}
\usepackage{color}
\usepackage{colortbl}
\usepackage{cuted}

\newcommand{\bv}[1]{\textit{\textbf{#1}}}

\newcommand{\din}{d_\text{in}}
\newcommand{\dout}{d_\text{out}}
\newcommand\numberthis{\addtocounter{equation}{1}\tag{\theequation}}
\title{One-Shot Learning as Instruction Data Prospector\\for Large Language Models}

\author{Yunshui Li$^{1,2}$\footnotemark[2] \quad  Binyuan Hui$^{3}$ \quad Xiaobo Xia$^{4}$ \quad Jiaxi Yang$^{1,2}$ \quad \textbf{Min Yang}$^{1}$\footnotemark[1]   \\ 
\textbf{Lei Zhang}$^{1,2}$ \quad \textbf{Shuzheng Si} \quad\textbf{Ling-Hao Chen} \quad \textbf{Junhao Liu} \\
\textbf{Tongliang Liu}$^4$  \quad \textbf{Fei Huang}$^{3}$ \ \  \textbf{Yongbin Li}$^{3}$\footnotemark[1] \\
        $^{1}$Shenzhen Institute of Advanced Technology, Chinese Academy of Sciences \\
        $^{2}$University of Chinese Academy of Sciences \\
        $^{3}$Alibaba Group \  $^{4}$The University of Sydney\\
        \texttt{\{ys.li, min.yang\}@siat.ac.cn, binyuan.hby@alibaba-inc.com} \\
        }

\begin{document}
\maketitle
\renewcommand{\thefootnote}{\fnsymbol{footnote}}
\footnotetext[1]{Corresponding authors}
\footnotetext[2]{Work done during the internship at Alibaba Group.}
\footnotetext[0]{\small\url{https://github.com/pldlgb/nuggets}}
\renewcommand{\thefootnote}{\arabic{footnote}}
\begin{abstract} 
Contemporary practices in instruction tuning often hinge on enlarging data scaling without a clear strategy for ensuring data quality, inadvertently introducing noise that may compromise model performance.
To address this challenge, we introduce \textsc{Nuggets}, a novel and efficient methodology that leverages one-shot learning to discern and select high-quality instruction data from extensive datasets. \textsc{Nuggets} assesses the potential of individual instruction examples to act as effective one-shot learning instances, thereby identifying those that can significantly improve performance across diverse tasks. 
\textsc{Nuggets} utilizes a scoring system based on the impact of candidate examples on the perplexity of a diverse anchor set, facilitating the selection of the most advantageous data for instruction tuning. 
Through comprehensive evaluations on two benchmarks, including MT-Bench and Alpaca-Eval, we show that instruction tuning with the top 1\% of examples curated by \textsc{Nuggets}
substantially outperforms conventional methods employing the entire dataset. 
\end{abstract}

\section{Introduction}

Large language models (LLMs)~\cite{Brown2020LanguageMA, OpenAI2023GPT4TR, Anil2023PaLM2T,bai2023qwen,li2023camel} 
have showcased remarkable capabilities~\cite{Wei2022EmergentAO, Schaeffer2023AreEA, Liu2023DoEA, cheng2024towards} across a wide range of language tasks by scaling the model size and training data.
Despite their proficiency, it is imperative to further enhance their alignment with human instructions.
This alignment process involves supervised fine-tuning (SFT) on input-output pairs, known as \textit{instruction tuning}. Instruction tuning is a crucial step, serving not only to activate the valuable knowledge acquired by LLMs during pre-training but also to facilitate their interaction with humans in a manner that aligns with natural conversational dynamics.

\begin{figure*}[t!]
    \centering
    \includegraphics[width=0.85\textwidth]{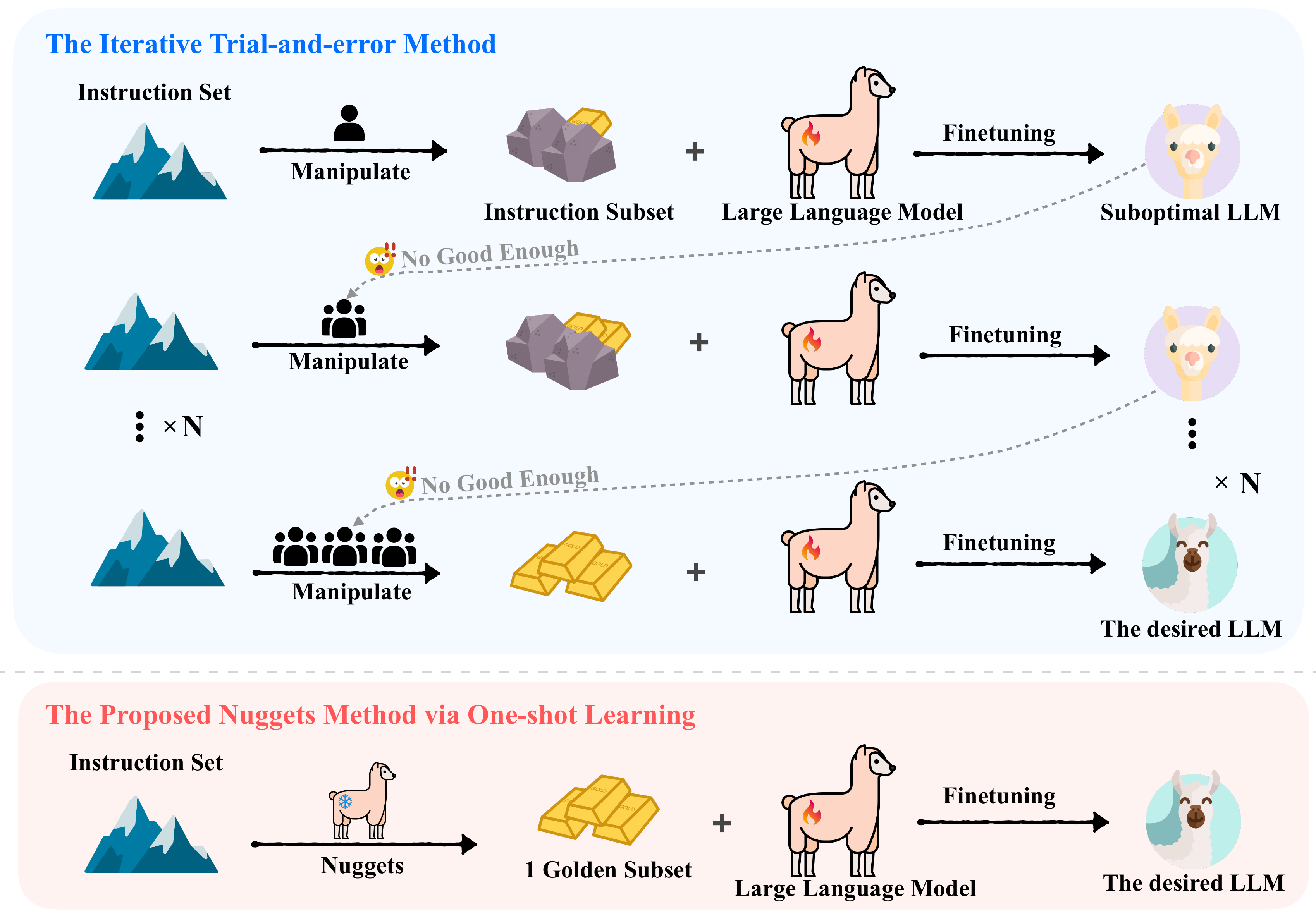}
    \caption{The comparison between our \textsc{Nuggets} and previous empirical methods. In contrast to empirical methods~(blue area), \textsc{Nuggets}~(orange area) can directly sample a gold subset, offering a more direct contribution to model fine-tuning. 
    }
    \label{fig:model}
\end{figure*}

Considerable efforts in instruction tuning have been concentrated on collecting larger ~\cite{chung2022scaling,wang2022super},
more diverse~\cite{sanh2022multitask,sun2023principle,wang2023far},  
and intricate~\cite{xu2023wizardlm, wei2023magicoder} datasets. This is commonly achieved through human crowd-sourcing~\cite{aghajanyan2021muppet,ouyang2022training,tang2022mvp}
or extracting data from larger pre-existing models~\cite{wang2022self,alpaca,vicuna2023, xu2023wizardlm}. 
Despite the growth in the size of datasets employed for instruction tuning, certain studies \cite{zhou2023lima, chen2023alpagasus,cao2023instruction} suggest that smaller yet valuable datasets tend to be more effective in harnessing the capabilities of LLMs.  Blindly expanding the volume of instruction data without ensuring quality may introduce noise and lead to hallucination issues~\cite{zhang2023siren,zhao2023survey}. However, there is a lack of standard criteria for selecting high-quality instruction data~\cite{li2023finding, har2004coresets,xia2022moderate, zhang2023ideal}. As depicted in Figure~\ref{fig:model}, the common practice depends on empirical methods for data selection~\cite{xia2023coreset}, introducing bias in determining data combinations and adjusting based on outcomes. This trial-and-error approach elevates alignment costs for models. We posit that optimal instruction combinations are present within the extensive data available, yet an efficient and cost-effective identification method remains underexplored.

In this paper, we introduce \textsc{Nuggets}, a simple yet efficient method that harnesses LLMs as data explorers through one-shot (in-context) learning. This method facilitates extracting high-quality, valuable data from expansive instruction datasets. 
Intuitively, an instructional example holds value in training if it serves as an excellent one-shot demonstration for a specific task. 
If it can facilitate many tasks, it will be worth being treated as a prime data focus, i.e., \textit{"gold instruction"}.
Another noteworthy perspective arises from the observation that in-context learning~\cite{dai2022can,yang2023iterative,wang-etal-2023-label} employs prompting to implicitly fine-tune the model, while instruction tuning operates through gradient descent. Leveraging the performance of in-context learning offers a promising avenue to predict the effects of instruction tuning. Concretely, we first select a set that spans multiple tasks, designated as the anchor set, and the dataset of instructions to be optimized is identified as the candidate set. One example is sequentially chosen from the candidate set to act as a one-shot example for in-context learning. Subsequently, it is scored based on its impact on the perplexity of each anchor example. This scoring mechanism enables the inference of dependencies between anchor and candidate examples, providing a reference standard for data selection.

To evaluate the effectiveness of the proposed \textsc{Nuggets}, we conduct extensive evaluations on two widely recognized benchmarks, namely MT-Bench~\cite{zheng2023judging} and Alpaca-Eval~\cite{alpaca_eval}. We choose a popular and powerful LLM, LLaMA~\cite{touvron2023llama}, as our base model. 
Experimental findings demonstrate that the \textsc{Nuggets}' data filtering strategy engenders a significant improvement in comparison to vanilla fine-tuning approaches. 

We summarize our main contributions as follows:
\begin{itemize}
    \item We present \textsc{Nuggets}, a methodology designed to dynamically assess the quality of instructional examples by using LLMs themselves. \textsc{Nuggets} is expected to extract the most valuable data from a vast pool of instruction data for the purpose of fine-tuning.
    \item Fine-tuning LLMs with solely the top 1\% of highest-scoring instructional examples yields superior results than using the entire instruction dataset. This observation underscores the significance of prioritizing the quality and strategic composition of the training data over sheer volume.
    \item The results of extensive experiments substantiate our hypotheses regarding \textit{"golden instructions"}, indicating that the effectiveness of an instructional example is measured by its impact on the task generalization capability of the model following the fine-tuning process. This observation holds considerable promise, potentially providing valuable insights for future endeavors in data quality screening.    
\end{itemize} 

\section{Related Work}
\paragraph{Instruction Tuning}
Recent works have introduced a series of techniques that aim to refine large language models (LLMs), showcasing their ability to generalize effectively to instructions not encountered before. For instance, T5~\cite{raffel2020exploring} pioneered the initial effort of training various natural language processing (NLP) tasks in a unified text-to-text format. FLAN~\cite{wei2021finetuned} introduced the novel concept of \textit{instruction tuning}, aiming to improve zero-shot task performance by transforming NLP tasks into natural language instructions during model training. Furthermore, InstructGPT~\cite{ouyang2022training} handled a wide array of human-created instructions encompassing diverse forms and a broad range of task types tailored for real-world user scenarios. 
In the absence of the source code release for these notable projects by OpenAI, subsequent efforts by Alpaca~\cite{alpaca, peng2023instruction} and Vicuna~\cite{vicuna2023} were undertaken to explore open-domain instruction tuning, employing the open-source LLM LLaMA~\cite{touvron2023llama}.

\paragraph{Instruction Construction}
The fine-tuning instruction datasets by previous methods are often created manually or tailored to specific tasks. To alleviate the issue of extensive human annotations and manual data gathering, various semi-automated techniques have emerged. Self-Instruct~\cite{wang2022self} randomly selected a limited number of instances from the initial task pool and used them as 
demonstrations to guide a language model in generating new instructions, along with their corresponding input-output pairs. 
Evol-Instruct~\cite{xu2023wizardlm} adopted a progressive modification strategy for the original instructions, which facilitated precise control over the difficulty and complexity levels of the generated instructions.
Tree-Instruct~\cite{zhao2023preliminary}, in contrast to Self-Instruct or Evol-Instruct, guided LLMs by instructing them to append a specified number of new nodes to the semantic tree of an existing instruction rather than directly manipulating the text sequence. Conversely, certain investigations are oriented towards augmenting the performance of LLMs by leveraging a reduced yet higher-quality set of instruction examples.
LIMA~\cite{zhou2023lima} demonstrated remarkably strong performance by strategically selecting a thousand high-quality data points for learning.
InstructMining~\cite{cao2023instruction} 
introduced a collection of carefully chosen natural language indicators for evaluating the quality of instruction-following data. Notably, this approach necessitates the division of data into multiple bins. Consequently, it encounters limitations in assessing the quality of individual examples at a fine-grained level. 
INSTAG~\citep{lu2023instag} proposed an open-set instruction tagging method to identify the semantics and intentions of human instructions through tags, providing definitions and quantified analyses of instruction diversity and complexity.
Moreover, ALPAGASUS~\cite{chen2023alpagasus} utilized the capabilities of an external and powerful model, ChatGPT, to directly evaluate each example. Despite the proven efficacy of this approach, a notable limitation lies in its inability to account for the inherent variations present in each model subjected to fine-tuning. It predominantly relies on the predilections of  ChatGPT. 
Although~\citet{li2023quantity} proposed a self-guided method for selecting data in instruction tuning, it still requires preliminary fine-tuning of the model, introducing uncertainty into subsequent operations.

\section{\textsc{Nuggets}}
\paragraph{Motivation}
As illustrated in Figure~\ref{fig:model}, the conventional paradigm for enhancing instructional data in the fine-tuning process of large language models (LLMs) has predominantly relied on empirical methods. These methods encompass the application of heuristic rules, expert analysis, and iterative adjustments to the data guided by feedback on model performance. Notably, this trial-and-error approach imposes significant costs in terms of both human effort and computational resources.

Recent scholarly consensus suggests that instruction tuning significantly enhances the task generalization capabilities of pre-trained models across various specific tasks~\cite{longpre2023flan,zhang2023instruction,zhang2023gpt4roi,shu2023exploitability}.  In light of this, we posit the hypothesis of a \textit{golden instruction}: the efficacy of an instructional example is gauged by its influence on the task generalization capability of the model subsequent to the fine-tuning procedure. As the extent of improvement becomes more conspicuous, the instruction gravitates towards classification as ``golden instruction''.

According to this hypothesis, a straightforward method involves fine-tuning an independent model using an instruction example and then comparing the performance of the fine-tuned model with the base model on a predefined dataset containing multiple tasks. This process aims to discern whether the given example qualifies as a ``golden instruction''.
However, this method would lead to an impractical proliferation of fine-tuned models, equivalent to the number of distinct instructions. Furthermore, fine-tuning with only a single example may introduce unstable updates to the model's gradients, making it challenging to ascertain the genuine acquisition of the example. Motivated by the inherent duality between \textit{In-Context Learning}~(ICL) and gradient descent~\cite{dai2022can, aizerman1964theoretical,yang2023iterative,irie2022dual}, we ``fine-tune'' the instruction implicitly through one-shot learning, replacing the need for actually fine-tuning the model.
More information can be found in Discussion~\ref{sec:diss}.

\begin{figure*}[t]
    \centering
    \includegraphics[width=0.9\textwidth]{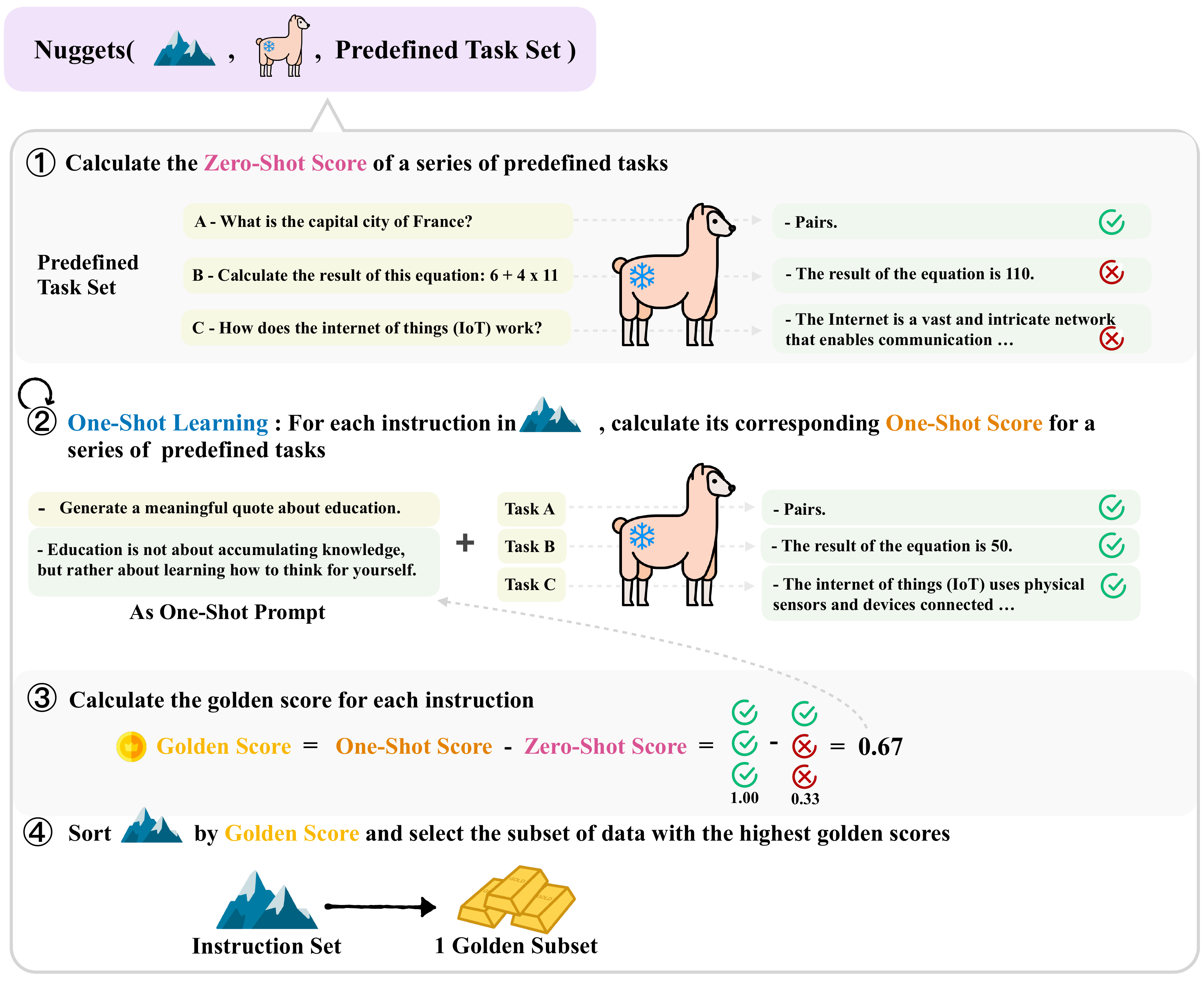}
    \caption{The illustration of the framework of our \textsc{Nuggets}. Note that we do not directly let the model generate answers for assessment. Instead, we calculate the model's logit scores on the ground truth answers as zero-shot scores or one-shot scores.}
    \label{fig:model2}
\end{figure*}

\paragraph{Overview}
The framework of our \textsc{Nuggets} is illustrated in Figure~\ref{fig:model2}.
Firstly, we evaluate the proficiency of LLMs across a diverse range of tasks using a predefined set of tasks, denoted as the \textit{zero-shot score}. Subsequently, we designate each example from the instruction dataset as a distinct \textit{one-shot prompt}, concatenating it in front of the predefined tasks.  
We then recalibrate the model's completion level for these tasks, referred to as the \textit{one-shot score}. 
By exploiting the disparity between one-shot and zero-shot scores, we can compute the golden score for each instruction. 
Once the golden scores for all instructions are computed, we can identify the highest-scoring subset, deemed the golden subset, which is subsequently provided directly to the model for the fine-tuning process.

\subsection{Algorithm Details}
\paragraph{Zero-Shot Score}

Given a predefined task set, it encompasses a variety of $m$ tasks, where each task is structured as $[\text{Task~(T)}, \text{Answer~(A)}]$. Each word in $\text{Task}$ or $\text{Answer}$ is denoted as ${w}^{\text{T}}_i$ or ${w}^{\text{A}}_i$. Let $\textsf{LLM}$  denote the pre-trained base large language model we use. For the $j$-th task that is represented by ${T}_j$, the probability of zero-shot inference by the model can be calculated by continuously predicting the next tokens based on the given task and the preceding words:
\begin{equation}
\begin{split}
        s^j_{\text{zsl}} &= \frac{1}{L}\sum_{i=1}^{L}\log p({w}^{\text{A}_j}_i | C;\textsf{LLM}), \\
        C &= [{T}_j, {w}_1^{\text{A}_j},{w}_2^{\text{A}_j},\ldots,{w}_{i-1}^{\text{A}_j}],
\end{split}
\end{equation}
where $L$ is the number of words of the ground-truth answer $\text{A}$. The score ${s}^j_{\text{zsl}}$ is employed to signify the extent of the model's proficiency on the $j$-th task. A higher ${s}^j_{\text{zsl}}$ denotes superior model performance on the $j$-th task, whereas a lower ${s}^j_{\text{zsl}}$ implies inferior performance. Therefore, we can acquire the model's performance across $m$ tasks as:
\begin{equation}
     \boldsymbol {S}_{\text{zsl}} = [{s}^1_{\text{zsl}}, {s}^2_{\text{zsl}},\ldots, {s}^{m-1}_{\text{zsl}}, {s}^{m}_{\text{zsl}}].
\end{equation}
\paragraph{One-Shot Score}

With an instruction tuning dataset $\mathcal{D}$, we aim to identify a set of examples $\mathcal{D}_{\text{gold}}$ that most closely align with the golden instructions. For each example $\mathbf{z}_k=[\text{Instruction}^{\text{Q}}_k~(\text{IQ}_k),\text{Instruction}^{\text{A}}_k~(\text{IA}_k)]$, we initially perform implicit instruction tuning on the base model using that specific example. 
Here, $\text{Instruction}^{\text{Q}}_k$ denotes the question associated with the $k$-th example $\mathbf{z}_k\in \mathcal{D}$, while $\text{Instruction}^{\text{A}}_k$ signifies its corresponding answer.
Subsequently, we employ the model with in-context learning to conduct another round of testing on the tasks within the predefined task set. That is,
\begin{equation}
    \begin{split}
        {s}^j_{\text{iit}}(\mathbf{z}_k)  
    &= \frac{1}{L}\sum_{i=1}^{L}\log p({w}^{\text{A}_j}_i |\underbrace{\text{IQ}_k,\text{IA}_k}_{\textit{One-Shot Prompt}}, C; \texttt{LLM}), \\
    C &=[{T}_j, {w}^{\text{A}_j}_1,{w}^{\text{A}_j}_2,\ldots,{w}^{\text{A}_j}_{i-1}], 
    \end{split}
    \label{eq3}
\end{equation}
where $\text{IQ}_k$ and $\text{IA}_k$ can be considered \textit{one-shot prompt}. 
Similarly, we can obtain the performance of the model after implicit fine-tuning across $m$ different tasks:

\begin{equation}
     \boldsymbol {S}^k_{\text{iit}} = [{s}^1_{\text{iit}}(\mathbf{z}_k), {s}^2_{\text{iit}}(\mathbf{z}_k),\ldots, {s}^{m-1}_{\text{iit}}(\mathbf{z}_k), {s}^m_{\text{iit}}(\mathbf{z}_k)].
\end{equation}
Afterward, we use the \textbf{Golden Score~(GS)} to reflect the impact of this instruction tuning example on the base model. The GS of the example $\mathbf{z}_k$ is calculated as 
\begin{equation}
    \text{GS}(\mathbf{z}_k) = \frac{1}{m}\sum_{i=1}^m \mathbbm{I}\left[{s}^i_{\text{iit}}(\mathbf{z}_k)> {s}^i_{\text{zsl}}\right]\  \in \  [0,1],
\end{equation}
where $\mathbbm{I}[\cdot]$ is the indicator function. At a high level, the GS measures the increment of performance improvement of the model after one-shot learning through the given instruction.

In this study, we calculate the GS score for each instructional example, facilitating the generation of a ranked list of scores encompassing the entire set of examples. Our objective is to explicitly fine-tune the base model by selectively employing a small subset comprising the most pivotal examples. Specifically, we prioritize examples exhibiting high golden scores, aiming to achieve superior outcomes compared to utilizing the entire dataset.
 
 \begin{table*}[t!]
    \centering
    \resizebox{0.87\textwidth}{!}{
    \begin{tabular}{l|r|cccccc|c}
    \toprule			
    \textbf{Model} & \textbf{Nums} & \textbf{Helpful\_Base} & \textbf{Koala} & \textbf{Self-instruct} & \textbf{Oasst} & \textbf{Vicuna} & \textbf{Length} & \textbf{Results}\\
    \midrule
    \midrule
    $\text{LLaMA}$ & - & 0.00 & 1.28 & 1.19  & 0.53  & 1.25 &2,980& 0.87 \\
    Alpaca$_{\text{full}}$ & 52,002 &20.15	& 25.64 & 27.77	& 25.00 & 15.00 & 396 & 25.43 \\
    \cmidrule(lr){1-9}
    Alpaca$_{\leq 0.5}$ & 9,542 & 7.75 & 5.12 & 13.09 & 9.57 & 8.75 & 241 & 10.96 \\
    Alpaca$_{>0.5}$ & 42,460 &24.03 & 20.51 & 28.57 & 29.78 & 15.00& 413 & 26.06 \\
    Alpaca$_{>0.8}$ & 7,525 &34.10 & 30.76 & 30.95 & 35.10 & 30.00 & 519 & \textbf{32.48} \\
    Alpaca$_{>0.85}$ & 619 & 37.20 & 26.90 & 25.00 & 29.30 & 22.50 &617 & \underline{28.20} \\
    \bottomrule
    \end{tabular}}
    \caption{The win\_rate results of various models under the Alpaca-Eval benchmark evaluation.}
    \label{alpaca-eval}
\end{table*}

\begin{table*}[t!]
    \centering
    \resizebox{0.9\textwidth}{!}{
    \begin{tabular}{l|cccccccc|c}
    \toprule				
    \textbf{Model} & \textbf{Writing}  & \textbf{Roleplay} & \textbf{Reasoning} & \textbf{Math} & \textbf{Coding} & \textbf{Extraction} & \textbf{STEM} & \textbf{Humanities}& \textbf{Overall}\\
    \midrule
    \midrule
    $\text{LLaMA}$ &4.6&	4.5&	5.2&	1.0	&1.20	&2.2&	5.0	&4.1&	3.47 \\
    Alpaca$_{\text{full}}$ & 8.5	&5.8	&3.3	&1.0	&2.0	&4.5	&6.5	&7.1	& 4.83 \\
    \cmidrule(lr){1-10}
    Alpaca$_{\leq 0.5}$ &7.2	&5.1	&2.1	&1.3	&1.9	&5.5	&5.3	&6.9	& 4.41 \\
    Alpaca$_{>0.5}$ & 8.3	&5.7	&3.5	&1.1	&1.7	&5.0	&6.6	&7	& 4.86\\
    Alpaca$_{>0.8}$ & 8.3	&5.9	&5.6	&1.8	&2.5	&4.0	&7.3	&7.4	&\textbf{5.34} \\
    Alpaca$_{>0.85}$ & 6.6	&6.3	&4.9	&1.0	&2.3	&3.3	&6.3	&7.3	&\underline{4.87} \\
    \bottomrule
    \end{tabular}}
    \caption{Experimental results of various models on the GPT-4 labeled MT-Bench benchmark.}
    \label{Mt-bench}
\end{table*}
\section{Experiments}
\subsection{Experimental Setup}
\paragraph{Instruction Dataset} 
We adopt the Alpaca dataset~\cite{alpaca} as instruction data. It is an important resource in the open-source community for instruction tuning, which is constructed by employing the self-instruct~\cite{wang2022self} method to distill instruction data from text-davinci-003. The success of this dataset in fine-tuning the LLaMA model has sparked a series of explorations into instruction fine-tuning~\cite{li2023m,ji2023exploring,xu2023baize}.
Besides, we perform more types of instruction datasets to verify the transferability of \textsc{Nuggets}, please refer to \ref{appendix_c}.

\paragraph{Predefined Task Set} 
The predefined task set plays a crucial role in computing golden scores for instructions. These data are employed to evaluate the model's ability to generalize across diverse tasks. The adequacy of the predefined task set is contingent upon its encompassing a substantial volume of data and incorporating a broad range of tasks. As the Alpaca dataset inherently possesses these attributes, we randomly choose 1,000 examples from it to constitute the predefined task set.

\paragraph{Evaluation Datasets}  
This work uses two methods to assess the model's capabilities. The first approach involves rating the responses generated by models on a scale ranging from 1 to 10. 
For this purpose, we utilize the GPT-4 labeled
\textbf{MT-Bench}~\cite{zheng2023judging} dataset, which evaluates instruction-following proficiency across eight categories: writing, roleplay, extraction, reasoning, math, coding, STEM, and humanities. Notably, since we only fine-tune on single-turn instruction data, the evaluation is restricted to \textit{Turn 1} of MT-Bench, similar to previous studies~\cite{cao2023instruction,zheng2023judging,chen2023alpagasus}. The second method involves comparing the model's generated responses with those produced by the Davinci-003 model, employing the well-established \textbf{Alpaca-Eval} dataset~\cite{alpaca_eval}. This dataset adopts the ``win\_rate'' as the evaluation metric.

\paragraph{Implementation Details}
 In our experiments, we designate the LLaMA-7B model as the foundational model. To ensure a fair comparison, we also set the maximum input length for the models fine-tuned with the Alpaca dataset to be consistent with LLaMA, which is 2048.  In the model fine-tuning phase, we employ the Adam optimizer with a learning rate of $2\times 10^{-5}$ and utilize a batch size of 64, conducting training over three epochs. In the subsequent model evaluation phase, we maintain all parameter settings consistent with the original work~\cite{alpaca_eval, zheng2023judging}.

\subsection{Experimental Results}
 The Alpaca dataset comprises a total of 52,002 instruction examples, and the distribution of their golden scores is illustrated in Appendix \ref{appendix_b}. Among these examples, 42,460 instances exhibit a golden score surpassing 0.5. In addition, a subset of examples closely aligned with the golden instructions has been selected, specifically those attaining golden scores above 0.8 and 0.85. In particular, there are 7,525 examples with golden scores surpassing 0.8 and 619 examples with golden scores exceeding 0.85. Notably, the latter subset constitutes a mere \textbf{1\%} of the entire dataset.

We conduct instruction tuning on the LLaMA model using various subsets of examples distinguished by their golden scores: those with scores less than 0.5, greater than 0.5, greater than 0.8, greater than 0.85, and the complete dataset. The fine-tuned models are denoted as $\text{Alpaca}_{\leq 0.5}$, $\text{Alpaca}_{>0.5}$, $\text{Alpaca}_{>0.8}$, $\text{Alpaca}_{>0.85}$, and $\text{Alpaca}_{\text{full}}$, respectively.

\begin{figure*}[t]
    \centering
    \includegraphics[width=0.86\textwidth]{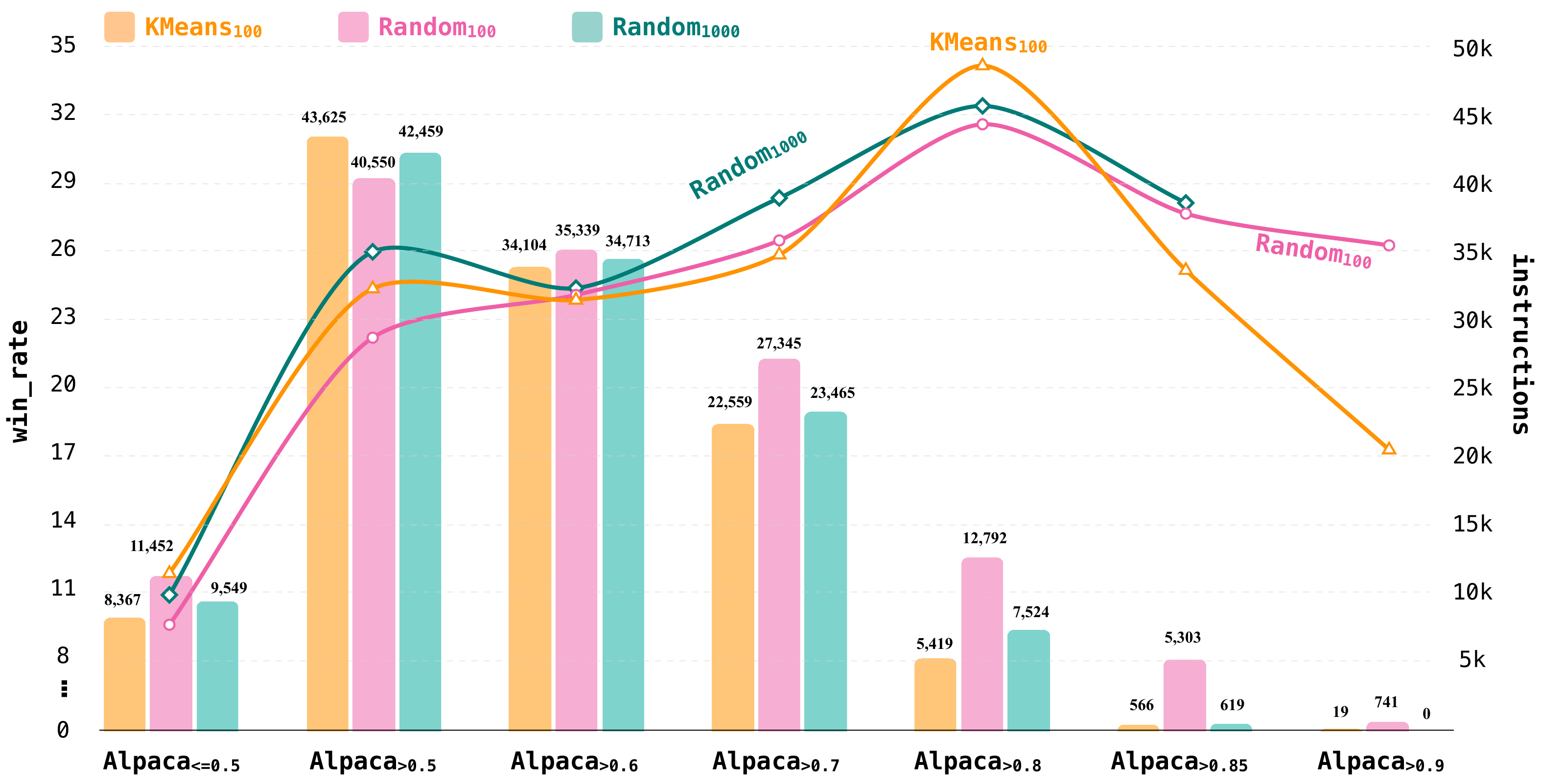}
    \caption{The distribution of the golden score for the instruction dataset across different predefined task sets, along with the corresponding fine-tuning results on the Alpaca-Eval benchmark.}
    \label{fig_ts}
\end{figure*}
\begin{table*}[!t]
    \centering
    \resizebox{0.85\textwidth}{!}{
    \begin{tabular}{lccccccc}
    \toprule			
    \textbf{Predefined Task Set} & \text{Alpaca}$_{\leq0.5}$ & \text{Alpaca}$_{> 0.5}$ & \text{Alpaca}$_{> 0.6}$ & \text{Alpaca}$_{> 0.7}$ & \text{Alpaca}$_{> 0.8}$& \text{Alpaca}$_{> 0.85}$& \text{Alpaca}$_{> 0.9}$\\
    \midrule
    \midrule
    K-Means$_{100}$ & 11.91& 24.44& 23.94& 25.93& 34.25	& 25.25 & 17.35\\
    \cmidrule(lr){1-8}
    Random$_{100}$ & 9.65& 22.28& 24.16& 26.56& 31.67	& 27.74 & 26.34\\
    \cmidrule(lr){1-8}
    Random$_{1000}$ & 10.96	& 26.06& 24.46& 28.43	& 32.48	& 28.21& - \\
    \bottomrule
    \end{tabular}}
    \caption{Win\_rate results on Alpaca-Eval Benchmark across different predefined task sets}
    \label{testselect}
\end{table*}
\paragraph{Main Results}
The experimental results are presented in Table~\ref{alpaca-eval} and Table~\ref{Mt-bench} for the Alpaca-Eval and MT-Bench benchmarks, respectively. As expected, $\text{Alpaca}_{>0.8}$ produces the most impressive outcomes. This can be attributed to its ability to maintain an optimal balance between the volume and quality of the instructions it utilizes, leading to the most desirable results. We also note that incorporating lower-quality instructions adversely affected model fine-tuning. This trend is clear when we see that $\text{Alpaca}_{\leq 0.5}$ lagged behind $\text{Alpaca}_{\text{full}}$ in performance, while $\text{Alpaca}_{>0.5}$ shows a slight edge over $\text{Alpaca}_{\text{full}}$. Remarkably, $\text{Alpaca}_{>0.85}$, using only 1\% of the dataset for fine-tuning, achieved results comparable to or even surpassing those of $\text{Alpaca}_{\text{full}}$. This underscores the efficacy of our data selection method.
More qualitative results can be found in Appendix~\ref{appendix_d}.

\begin{figure*}[h]
    \centering
    \includegraphics[width=0.87\textwidth]{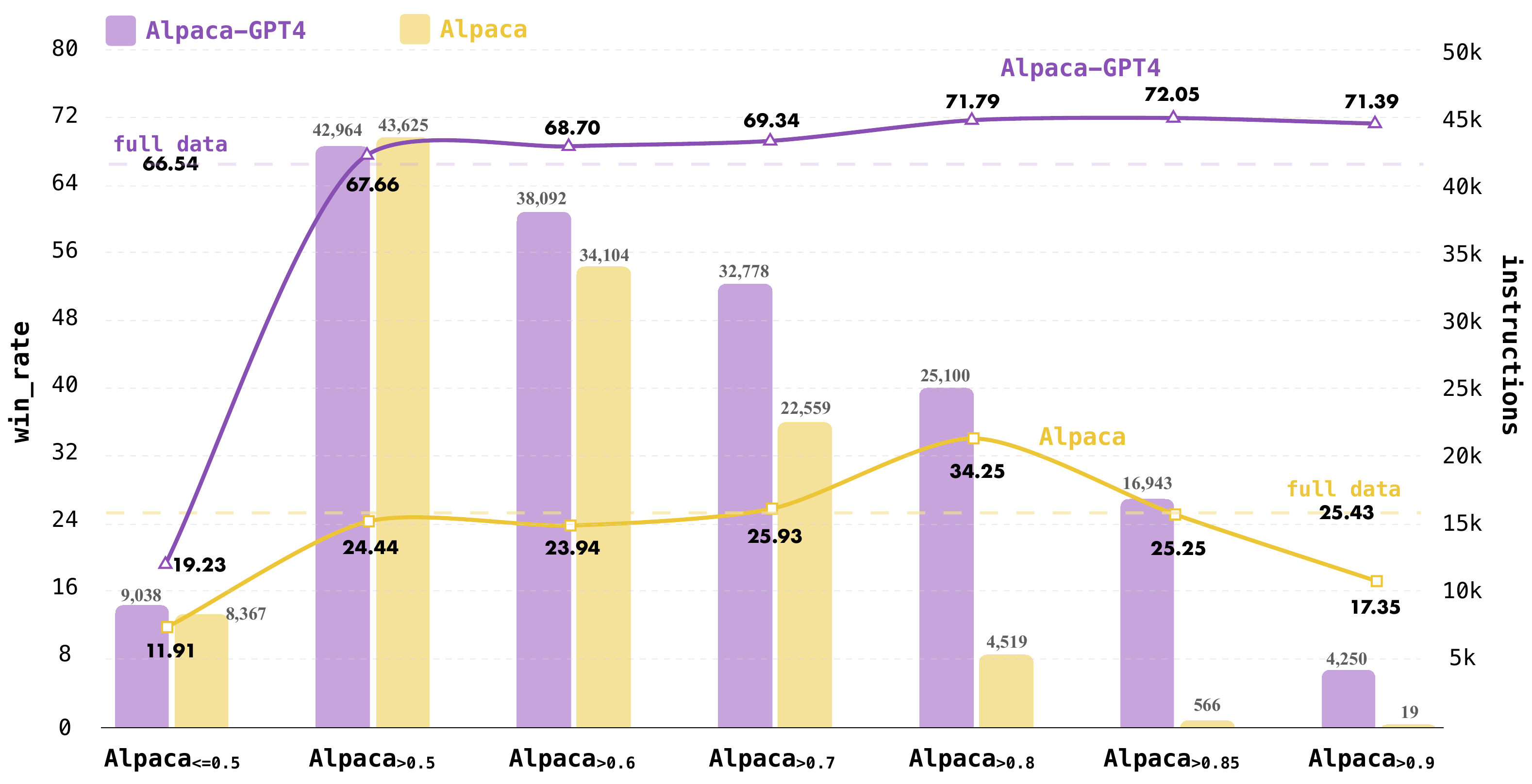}
    \caption{The distribution of the golden score for the instruction dataset across different instruction sets, along with the corresponding fine-tuning results on the Alpaca-Eval benchmark. Both predefined task sets utilize K-Means to sample 100 examples from their respective instruction datasets.}
    \label{fig_alpaca_gpt4}
\end{figure*}

\begin{table*}[h]
    \centering
    \resizebox{0.85\textwidth}{!}{
    \begin{tabular}{llcccccccc}
    \toprule			
    \textbf{\ } & & \textbf{GS}$_{\leq0.5}$ & \textbf{GS}$_{> 0.5}$ & \textbf{GS}$_{> 0.6}$ & \textbf{GS}$_{> 0.7}$ & \textbf{GS}$_{> 0.8}$& \textbf{GS}$_{> 0.85}$& \textbf{GS}$_{> 0.9}$ & \textbf{Full Data}\\
    \midrule
    \midrule
    \multirow{2}{*}{\textbf{LLaMA2}} &NUM& 3,730 & 48,272 & 40,905 & 28,644 & 10,409 & 2,411 & 87  &52,002 \\
    & Win\_rate & 13.17 & 27.09 & 27.85 & 27.62 & \underline{33.92} & \textbf{34.98} &27.08 & 26.47 \\
    \midrule
    \multirow{2}{*}{\textbf{Mistral}} &NUM& 78 & 51,924 & 51,610 & 49,398 & 36,068 & 23,147 & 9,356 & 52,002  \\
    & Win\_rate & 0 & 12.26 & 11.10 & \underline{12.45} & 11.28 & 10.60 & \textbf{13.53} & 9.85  \\

    \bottomrule
    \end{tabular}}
    \caption{Win\_rate results on Alpaca-Eval Benchmark across two different foundation models.}
    \label{foundation-model}
\end{table*}
\paragraph{Ablation on Predefined Task Sets}
To evaluate how different predefined task sets affect the selection of instruction data for fine-tuning, we include two additional predefined task set variations. One is randomly exampled from the Alpaca dataset but with a smaller task set size, which is limited to 100 examples. The other one entails clustering the Alpaca dataset into 100 clusters using the K-Means algorithm and selecting the centroids of each cluster as examples of the task set. 

We use the two predefined sets to calculate golden scores for the Alpaca dataset separately. The distribution of golden scores is depicted in Figure~\ref{fig_ts}. We select instruction data with golden scores less than or equal to 0.5, greater than 0.5, greater than 0.6, greater than 0.7, greater than 0.8, greater than 0.85, and greater than 0.9 for model fine-tuning, respectively. Table~\ref{testselect} suggests that with random sampling, increasing the size of the task set can enhance the identification of high-quality instruction data. The logic behind this is that a larger encompasses a broader diversity of data, facilitating a more nuanced assessment of an instruction's effect on model task generalization. However, a shift occurs when K-Means is employed to cherry-pick more distinct examples for the task set. With as few as 100 examples, K-Means outshines the results from 1,000 examples acquired through random sampling. In this instance, $\text{Alpaca}_{> 0.8}$ delivered a superior performance with just 5,419 examples, compared to the 7,524 examples seen with $\text{Random}_{1000}$. This outcome also indirectly confirms the validity of our hypothesis regarding the definition of golden instructions.

\paragraph{Ablation on Instruction Sets} 
To delve deeper into the generalization capabilities of \textsc{Nuggets} across varied instruction datasets, we undertake a series of experiments utilizing the Alpaca-GPT4 dataset~\cite{peng2023instruction}. It generates instructional data from the powerful GPT-4 model~\cite{OpenAI2023GPT4TR}, which is considered to have superior data quality. Additionally, it shares the same questions in instructions with the Alpaca dataset, which facilitates our direct comparison between the two. 

Inspired by Table~\ref{testselect}, we employ the K-Means algorithm on the Alpaca-GPT4 dataset to sample 100 examples, forming the predefined task set. Subsequently, we apply the \textsc{Nuggets} method to score all instructions in the dataset with the golden score, as depicted in Figure~\ref{fig_alpaca_gpt4}. Compared to the Alpaca dataset, the Alpaca-GPT4 dataset boasts a higher number of instructions with golden scores: 25,100 instructions exceed a score of 0.8, 16,943 surpass 0.85, and 4,250 instructions exceed 0.9. These numbers far exceed the corresponding high-scoring instructions in the Alpaca dataset. 
This also demonstrates that the golden score can serve as an absolute metric to assess the quality of instructional data.
The results from model fine-tuning indicate that on the Alpaca-GPT4 dataset, conclusions align with those of previous experiments. The large language models fine-tuned on subsets with golden scores less than or equal to 0.5 exhibit the poorest performance, with a win rate of only 19.23\% in the Alpaca-Eval benchmark. In contrast, the models fine-tuned on subsets with golden scores greater than 0.85 demonstrate superior performance, boasting a high win rate of 72.05\%. This success can be attributed to the dual assurance of quantity and quality in this particular subset. It is worth emphasizing that fine-tuning on a small and high-quality dataset consistently and significantly outperforms the results of fine-tuning on the full dataset. Overall, the models fine-tuned using Alpaca-GPT4 significantly outperform those fine-tuned with Alpaca. This implicitly corroborates the superior quality of the Alpaca-GPT4 dataset compared to the Alpaca dataset. For more experiments on instruction datasets, please refer to Appendix~\ref{appendix_c}.
\paragraph{Ablation on Foundation Models}
To verify the transferability of the \textsc{Nuggets} method, we conducted experiments on different foundation models using the Alpaca instruction dataset.
We selected LLaMA2~\citep{touvron2023llama2} and Mistral~\citep{jiang2023mistral} at the 7B size as the new base models. 
The distribution of the golden scores and the performance of models fine-tuned on corresponding subsets of instructions are shown in Table~\ref{foundation-model}.
We found that the \textsc{Nuggets} method is also applicable to other models. LLaMA2 achieved the best results under fine-tuning on subsets with a golden score greater than 0.85, reaching 34.98, which is significantly higher than the 26.47 achieved under full data. Although the absolute value of the win\_rate for the Mistral series of fine-tuned models is somewhat low, their performance is also significantly boosted by the \textsc{Nuggets} data filtering.

\section{Discussion: One-Shot Learning as Implicit Instruction Tuning}
\label{sec:diss}
Transformer has risen as the prevailing architecture for language models, where self-attention plays a crucial role as a pivotal element within Transformer. Let $\mathbf{X_{\text{ins}}}, \mathbf{X_{\text{test}}} \in \mathbb{R}^{\din} $ denote the instruction tuning sample and the test input respectively. 
 $\mathbf{X}_{\text{ins}}$ can be likened to $\text{IQ}_k$ and $\text{IA}_k$ in Equation~\ref{eq3}, while $\mathbf{X}_{\text{test}}$ can be seen as $\text{T}$ and ${w}^{\text{A}}_1,{w}^{\text{A}}_2,\ldots,{w}^{\text{A}}_{i-1}$. That $\bv{Q} = \mathbf{W}_Q \mathbf{X_{\text{test}}^\top} $ be the attention query vector, $\bv{K} = \mathbf{W}_K [\mathbf{X}_{\text{ins}} \Vert \mathbf{X}_{\text{test}} ]$ be the attention key vector and $\bv{V} = \mathbf{W}_V [\mathbf{X}_{\text{ins}} \Vert \mathbf{X}_{\text{test}} ]$ be the attention value vector, where $\Vert $ represents concatenation operation, $\mathbf{W}_K, \mathbf{W}_V, \mathbf{W}_Q \in \mathbb{R}^{\dout \times \din}$ are the projection matrices for computing the attention queries, keys, and values, respectively. The result of self-attention in an arbitrary layer for a head is formulated as:

\begin{figure}[h]
\small
\begin{align*}
    & \textsf{Attention}(\bv{K}, \bv{V}, \bv{Q}) \\
    &= \mathbf{W}_V [\mathbf{X}_{\text{ins}} \Vert \mathbf{X}_{\text{test}} ]  \textsf{Softmax} \left(\frac{\mathbf{W}_K [\mathbf{X}_{\text{ins}} \Vert \mathbf{X}_{\text{test}} ] ^\top \bv{Q}} {\sqrt{ \din }}\right)\\
    &\approx \mathbf{W}_V [\mathbf{X}_{\text{ins}} \Vert \mathbf{X}_{\text{test}} ] \left(\mathbf{W}_K [\mathbf{X}_{\text{ins}} \Vert \mathbf{X}_{\text{test}}]\right) ^\top \bv{Q} \\
    &= \underbrace{\mathbf{W}_V\mathbf{X}_{\text{test}}(\mathbf{W}_K\mathbf{X}_{\text{test}})^\top}_{\textit{ Only test input.}} \bv{Q} + \underbrace{\mathbf{W}_V\mathbf{X}_{\text{ins}}(\mathbf{W}_K\mathbf{X}_{\text{ins}})^\top}_{\textit{Only instruction sample.}} \bv{Q} \\
&= \mathbf{W}_\text{zsl}\ \bv{Q} + \Delta \mathbf{W}_\text{iit}\  \bv{Q} \\
&= (\mathbf{W}_\text{zsl} + \Delta \mathbf{W}_\text{iit})\  \bv{Q},  \numberthis
\end{align*}
\end{figure}
where $\sqrt{d_{\text{in}}}$ serves as a scaling factor. The term $\mathbf{W}_V\mathbf{X}_{\text{test}}(\mathbf{W}_K\mathbf{X}_{\text{test}})^\top$ could be denoted as $\mathbf{W}_\text{zsl}$, which represents the zero-shot learning scenario where no instruction tuning is performed since it solely focuses on the test input. In addition,
the term $\mathbf{W}_V\mathbf{X}_{\text{ins}}(\mathbf{W}_K\mathbf{X}_{\text{ins}})^\top$ can be seen as implicit instruction tuning $\Delta \mathbf{W}_\text{iit}$ achieved via the meta-gradient \cite{dai2022can, yang2023iterative} derived from the instruction sample. Readers can refer to previous papers~\cite{dai2022can, aizerman1964theoretical,irie2022dual} for more details on implicit instruction tuning.

\section{Conclusion} 

This paper presents \textsc{Nuggets}, a method leveraging LLMs to discern more pivotal data for instruction tuning. Grounded in one-shot learning, this approach facilitates the identification of examples' value, enabling efficient data selection without dependence on additional annotation and associated costs. Benefiting from \textsc{Nuggets}, we observe improved instruction following abilities even with smaller training subsets. Furthermore, we posit that our method underscores the significance of meticulous data selection, offering valuable insights for future instruction fine-tuning endeavors. 

\section*{Limitations} 
Although the efficacy of the proposed approach has been confirmed through empirical experiments, opportunities for refinement persist. One avenue for improvement involves a thorough investigation into the inclusion of a diverse and compact set of \textit{predefined tasks} during the golden scoring phase. This exploration aims to enhance the efficiency of model evaluation on instructional data, leading to improved identification of high-quality instructions suitable for subsequent model fine-tuning. 
Secondly, due to resource constraints, the majority of experiments in this study are confined to the LLaMA-7B model. While this model holds significant influence within the large language model open-source community, comprehensive validation across a broader spectrum of models is imperative to ensure the generalizability of the proposed approach.
Lastly, to fortify the empirical foundation of our findings, it is crucial to subject the proposed method to validation on a more extensive array of instructional datasets. This step aims to ascertain the robustness and applicability of the methodology across a diverse range of instructional contexts, contributing to its broader utility in real-world scenarios. These outlined avenues for future work are anticipated to refine and extend the scope of our proposed method.

\section*{Acknowledgments}
Min Yang was supported by National Key Research and Development Program of China (2022YFF0902100), National Natural Science Foundation of China (62376262), the Natural Science Foundation of Guangdong Province of China (2024A1515030166), Shenzhen Science and Technology Innovation Program (KQTD20190929172835662), Shenzhen Basic Research Foundation (JCYJ20210324115614039). This work was supported by Alibaba Group through Alibaba Innovative Research Program.
Xiaobo Xia was supported by the Australian Research Council project: DE190101473 and Google
PhD Fellowship. Tongliang Liu is partially supported by the following Australian Research Council projects: FT220100318, DP220102121, LP220100527, LP220200949, and IC190100031.

\clearpage 

\bibliography{custom}
\clearpage
\onecolumn

\appendix

\section{The Distribution of Golden Score}
\label{appendix_b}
\begin{figure*}[h]
    \centering
    \includegraphics[width=0.8\textwidth]{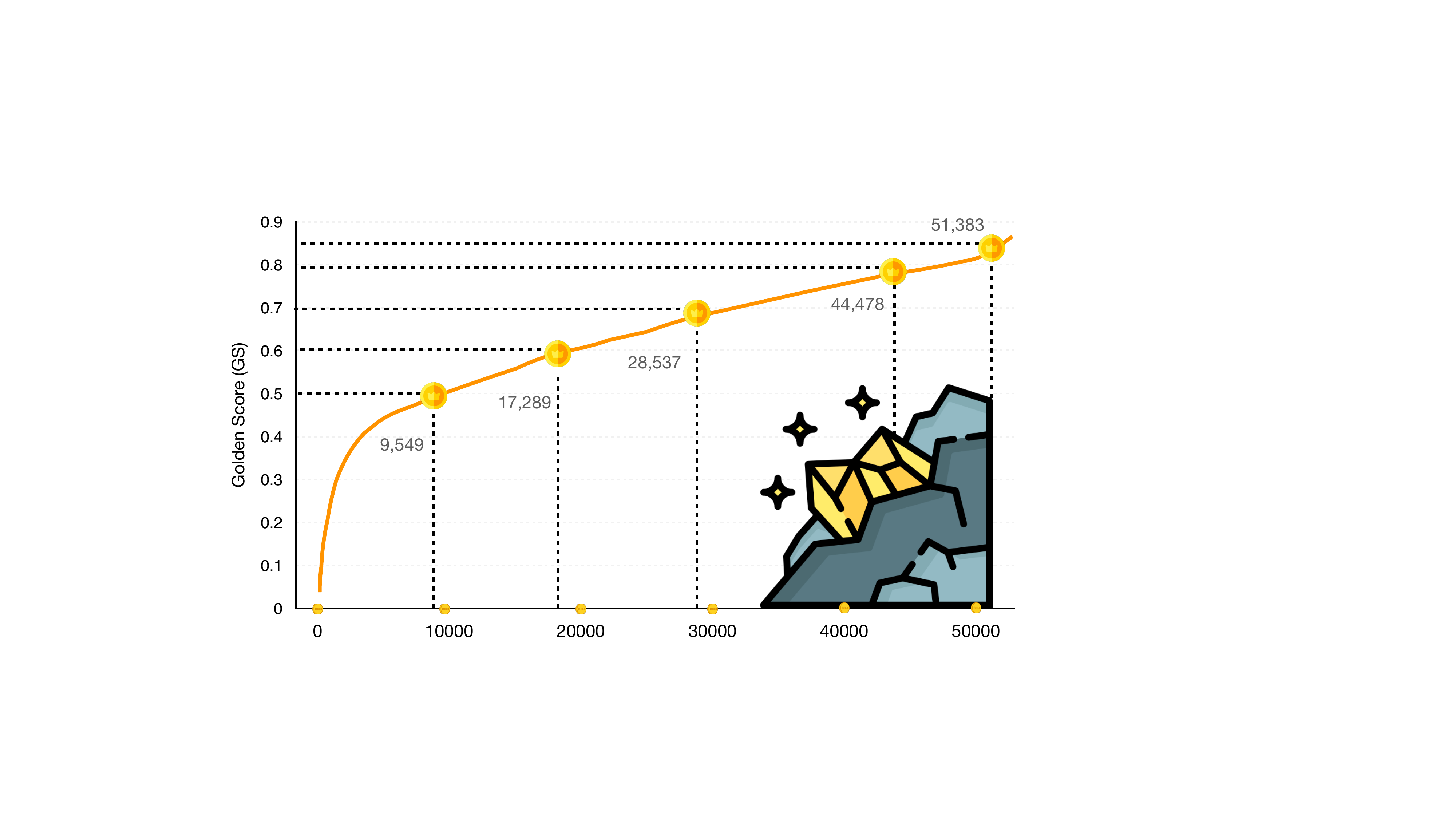}
    \caption{The distribution of the golden score for the Alpaca instruction dataset. }
    \label{fig_gs}
\end{figure*}

As shown in Figure~\ref{fig_gs}, in a total of 52,002 cases, there are 9,549 instructions with a gold score of less than 0.5, indicating that these data have a side effect on overall task completion. Besides, there are 7,524 instructions with a gold score greater than 0.8, suggesting that the model improves the task completion rate through one-shot learning from these data, which can be considered high-quality instruction data.
\section{Experiment on Other Instruction Sets}
\label{appendix_c}
Based on the LLaMA-7B model, we conducted experiments on several other instruction datasets, further validating the effectiveness of our \textsc{Nuggets} method.
\subsection{Code Alpaca}
 The Code Alpaca instruction dataset~\citep{codealpaca} is designed to develop large language models capable of following instructions and generating code. Leveraging self-instruct~\citep{wang2022self} technology, it has produced 20,000 examples of instruction data. 
 We use HumanEval~\citep{chen2021evaluating} as a benchmark to evaluate the model's code generation capabilities. It is used to measure functional correctness for synthesizing programs from docstrings. It consists of 164 original programming problems, assessing language comprehension, algorithms, and simple mathematics, with some being comparable to simple software interview questions. 
 We adopt the approach outlined by \citet{chen2021evaluating} to calculate pass rates at k values of 1, 10, and 100 for each problem. Essentially, pass@1 predicts the probability of a model producing a correct solution on the first try, while pass@10 and pass@100 predict the probability of achieving a correct solution within 10 and 100 tries, respectively. We generate 200 completions at a temperature setting of 0.2~\citep{luo2023wizardcoder} to estimate pass@1, pass@10, and pass@100 rates.
 \begin{figure*}[h]
    \centering
    \includegraphics[width=1.0\textwidth]{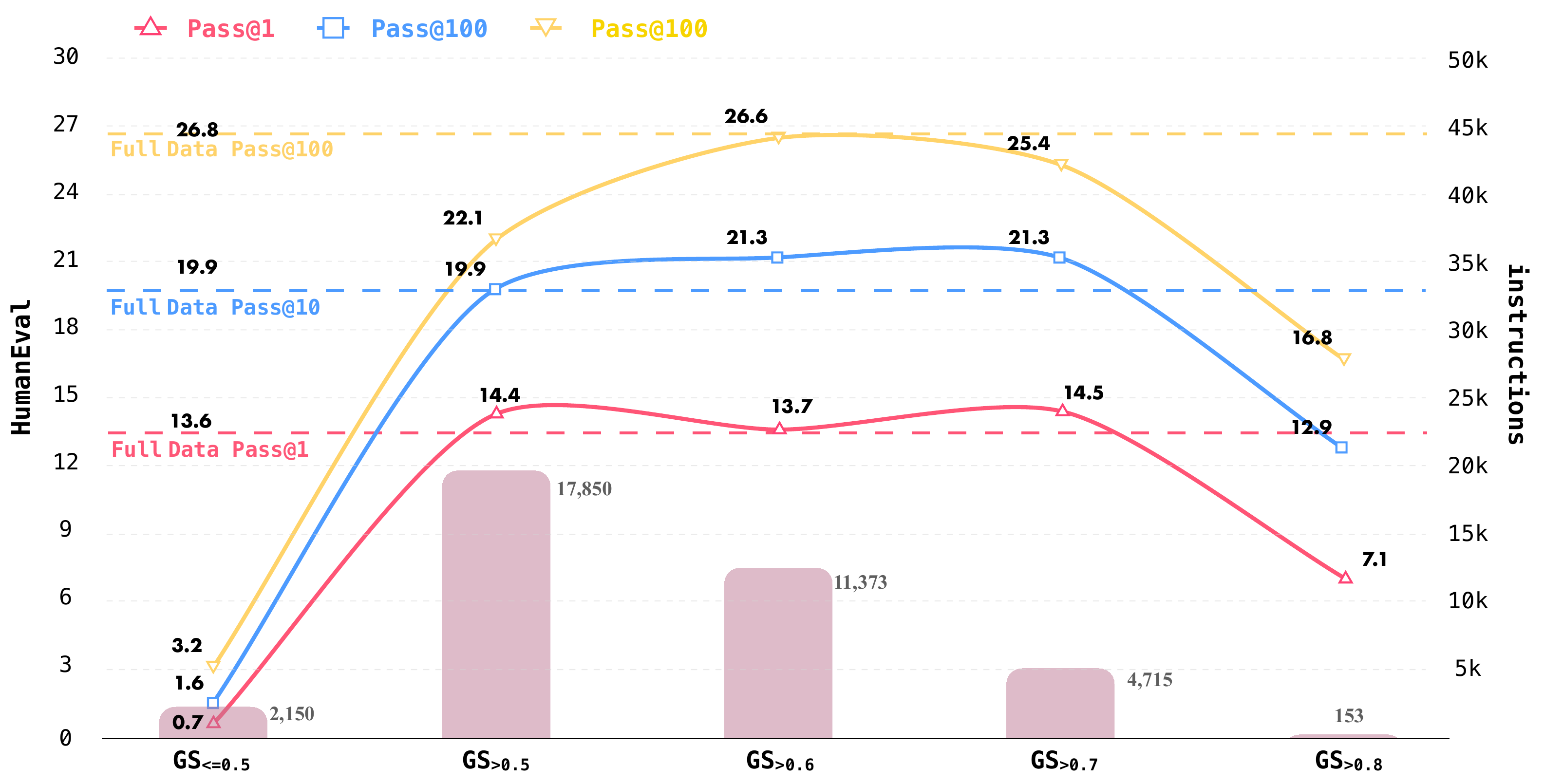}
    \caption{The distribution of the golden score for the Code Alpaca instruction dataset, along with the corresponding fine-tuning results on the HumanEval benchmark. Predefined task sets utilize K-Means to sample 100 examples from the Code Alpaca instruction dataset.}
    \label{code_alpaca}
\end{figure*}
The experimental results are shown in the Figure~\ref{code_alpaca}. Out of 20,000 instructions, 4,715 instructions have a gold score greater than 0.85, achieving the best pass@1 and pass@10 results in the HumanEval benchmark, superior to the fine-tuning results of the full dataset. Additionally, this experiment also proves that the \textsc{Nuggets} method can be applied to fine-tuning for specific tasks, demonstrating good transferability.

\subsection{WizardLM}
The WizardLM instruction dataset~\cite{xu2023wizardlm}, which employs Evol-Instruct to iteratively refine an initial set of instructions into more complex ones, contains 70,000 instruction examples. The distribution of the golden scores and the performance of models fine-tuned on corresponding subsets of instructions are shown in Table~\ref{wizard}. 
\begin{table*}[h]
    \centering
    \resizebox{1.0\textwidth}{!}{
    \begin{tabular}{lccccccccc}
    \toprule			
    \textbf{\ } & \textbf{GS}$_{\leq0.5}$ & \textbf{GS}$_{> 0.5}$ & \textbf{GS}$_{> 0.6}$ & \textbf{GS}$_{> 0.7}$ & \textbf{GS}$_{> 0.8}$& \textbf{GS}$_{> 0.85}$& \textbf{GS}$_{> 0.86}$& \textbf{GS}$_{> 0.87}$ & \textbf{Full Data}\\
    \midrule
    \textbf{NUM} & 480 & 69,520 & 69,377 & 68,898 & 65,190 & 40,223 & 23,579 &3, 316 & 70,000 \\
    
    \cmidrule(lr){1-10}
    \textbf{Win\_rate} & 19.42 & 58.08 & 57.40 & 56.21 & \textbf{59.81} & \underline{58.40} & 57.81 & 54.68 & 57.65 \\
    \bottomrule
    \end{tabular}}
    \caption{The distribution of golden scores for the WizardLM dataset and the evaluation results of models fine-tuned on corresponding score subsets on the Alpaca-Eval benchmark.}
    \label{wizard}
\end{table*}
We can observe that the quality distribution of the WizardLM dataset is relatively balanced, with 65,190 instruction examples having a golden score greater than 0.8, accounting for 93\% of the total number of instructions. In the evaluation of the Alpaca-Eval benchmark, models fine-tuned on subsets with golden scores greater than 0.8 achieved a win rate of 59.81, outperforming models fine-tuned on the full dataset.
\subsection{FLANv2}
We sampled 50,000 examples from the FLANv2~\citep{chung2022scaling} dataset to constitute the instruction tuning data for this experiment. Additionally, the Predefined task set was also derived from these 50,000 examples, using the K-Means algorithm to sample 100 examples. 
We evaluated the performance of the fine-tuned model using MMLU~\cite{hendrycks2021measuring} in a 5-shot setting. MMLU is a test designed to measure a text model's multitask accuracy. The test encompasses 57 tasks, including elementary mathematics, US history, computer science, law, and more. 
\begin{table*}[h]
    \centering
    \resizebox{0.9\textwidth}{!}{
    \begin{tabular}{lcccccccc}
    \toprule			
    \textbf{\ } & \textbf{GS}$_{\leq0.5}$ & \textbf{GS}$_{> 0.5}$ & \textbf{GS}$_{> 0.6}$ & \textbf{GS}$_{> 0.7}$ & \textbf{GS}$_{> 0.8}$& \textbf{GS}$_{> 0.85}$& \textbf{GS}$_{> 0.9}$ & \textbf{Full Data}\\
    \midrule
    \textbf{NUM} & 1,361 & 48,639 & 46,009 & 39,046 & 17,037 & 4,798 & 321  & 50,000 \\
    
    \cmidrule(lr){1-9}
    \textbf{Acc} & 34.68 & 41.38 & \underline{41.92} & 41.87 & \textbf{41.97} & 35.51 & 26.41 & 40.45 \\
    \bottomrule
    \end{tabular}}
    \caption{The distribution of golden scores for the sampled FLANv2 dataset and the evaluation results of models fine-tuned on corresponding score subsets on the MMLU benchmark.}
    \label{flan}
\end{table*}
The experimental results are shown in the table. It can be observed that the model fine-tuned with examples having a golden score greater than 0.8 (totaling 17,037 examples) achieved the best results, followed by those fine-tuned with examples having a golden score greater than 0.6 (totaling 46,009 examples). Additionally, we noted that the model fine-tuned with examples having a golden score less than 0.5 (totaling 1,361 examples) outperformed the model fine-tuned with examples having a golden score greater than 0.9 (totaling 321 examples).

We speculate that the reason for this outcome is that the MMLU evaluation itself is a knowledge-intensive assessment, which thus imposes higher demands on the quantity of fine-tuning examples. In the future, some optimizations in the selection of the Predefined Task Set may potentially improve this situation.
\section{Case Study}
\label{appendix_d}
\begin{figure*}[t]
    \centering
    \includegraphics[width=\textwidth]{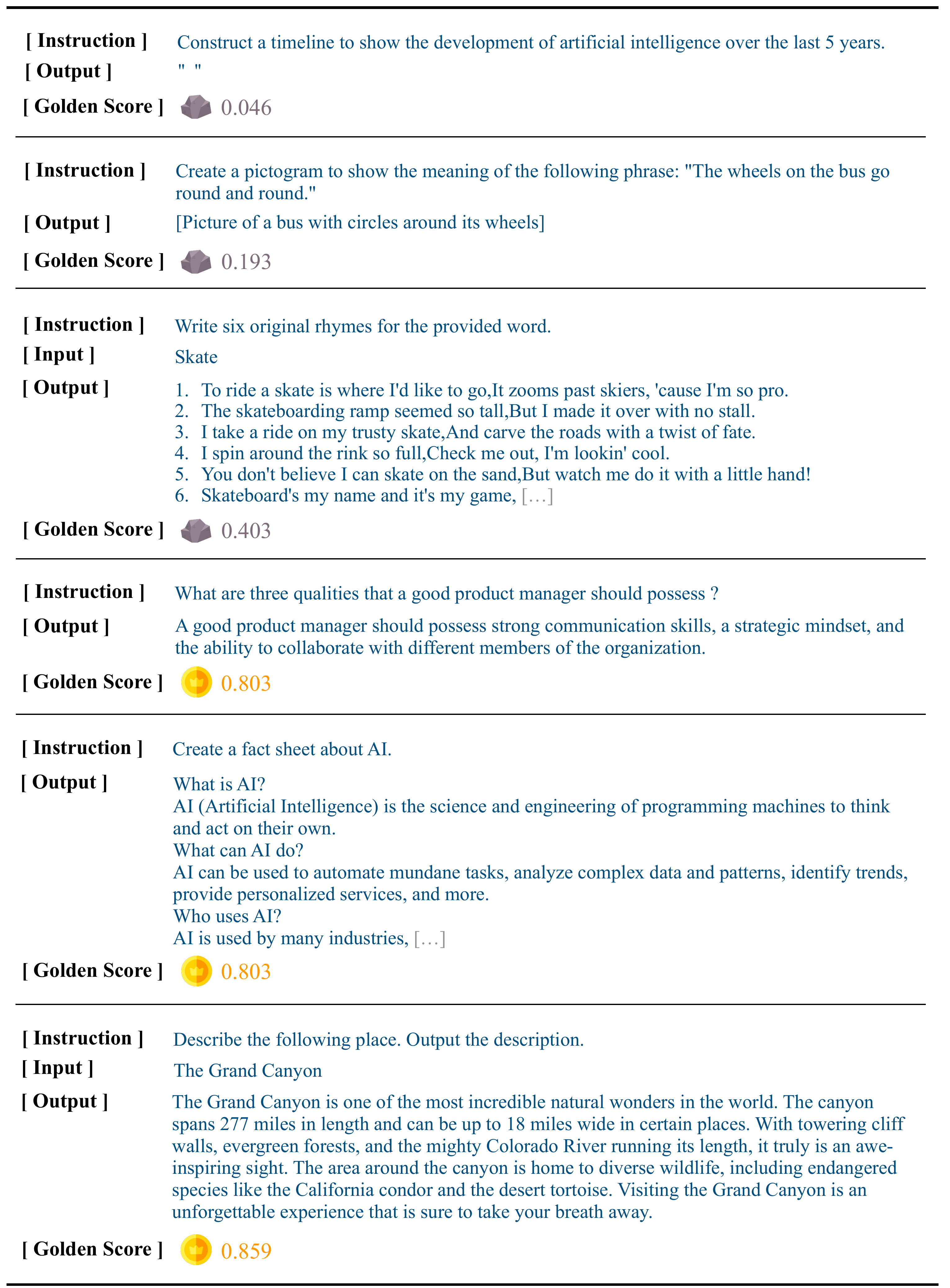}
    \caption{Examples of instructions and their corresponding golden scores.}
    \label{fig_case}
\end{figure*}

 To evaluate \textsc{Nuggets} qualitatively, we also select some instruction examples from the Alpaca dataset for case studies, as shown in Figure~\ref{fig_case}. We observe that noisy instructions without outputs display extremely low golden score values. Such examples consistently score very low throughout the instruction dataset. Additionally, instructions with hallucinatory outputs and some that are a bit confusing also typically exhibit lower golden scores. In contrast, instructions with higher scores often possess fluent language and meticulous logic, and are oriented towards assisting humans in problem-solving. Therefore, we believe that the high golden scores can represent high-quality instruction tuning data.  

\end{document}